\documentclass[sigconf]{acmart}

\usepackage{amsmath}
\usepackage{tipa}
\usepackage{multirow}
\usepackage{caption}
\usepackage{subcaption}
\usepackage{mdframed}
\usepackage{enumitem}

\AtBeginDocument{%
  \providecommand\BibTeX{{%
    \normalfont B\kern-0.5em{\scshape i\kern-0.25em b}\kern-0.8em\TeX}}}

\setcopyright{none}

\newcommand{\qq}[1]{``#1''}

\newcommand{\RQN}[2]{\textbf{RQ#1} \textit{#2}}
\newcommand{\ie}{i.e.,~}
\newcommand{\eg}{e.g.,~}
\newcommand{\cf}{cf.~}
\newcommand{\rNum}[1]{\expandafter{\romannumeral #1\relax}}

\begin{document}
\fancyhead{}

\title{Triple Classification for Scholarly Knowledge Graph Completion}

\author{Mohamad Yaser Jaradeh}
\affiliation{\institution{L3S Research Center, Leibniz University Hannover}
\city{Hanover}
\country{Germany}}
\email{jaradeh@l3s.de}
\orcid{0000-0001-8777-2780}

\author{Kuldeep Singh}
\affiliation{\institution{Cerence GmbH \and Zerotha Research
\city{Aachen}
\country{Germany}}}
\email{kuldeep.singh1@cerence.com}
\orcid{0000-0002-5054-9881}

\author{Markus Stocker}
\affiliation{\institution{TIB Leibniz Information Centre for Science and Technology
\city{Hanover}
\country{Germany}}}
\email{markus.stocker@tib.eu}
\orcid{0000-0001-5492-3212}

\author{S\"oren Auer}
\affiliation{\institution{TIB Leibniz Information Centre for Science and Technology
\city{Hanover}
\country{Germany}}}
\email{auer@tib.eu}
\orcid{0000-0002-0698-2864}

\renewcommand{\shortauthors}{Jaradeh et al.}

\begin{abstract}
Scholarly Knowledge Graphs (KGs) provide a rich source of structured information representing knowledge encoded in scientific publications. 
With the sheer volume of published scientific literature comprising a plethora of inhomogeneous entities and relations to describe scientific concepts, these KGs are inherently incomplete. 
We present exBERT (\textipa{"Eksp3:Rt}), a method for leveraging pre-trained transformer language models to perform scholarly knowledge graph completion. 
We model triples of a knowledge graph as text and perform triple classification (\ie belongs to KG or not).
The evaluation shows that exBERT outperforms other baselines on three scholarly KG completion datasets in the tasks of triple classification, link prediction, and relation prediction.
Furthermore, we present two scholarly datasets as resources for the research community, collected from public KGs and online resources.
\end{abstract}


\maketitle

\section{Introduction}
\label{sec:intro}

The proliferation of scientific literature creates new challenges in the research community, such as reproducibility crisis, duplication, inefficiency, and a lack of transparency \cite{white2017science}. Thus, organizing scholarly knowledge is one of the most pressing tasks for solving current and upcoming societal challenges. 

In general, Knowledge Graphs (KGs) have become central sources to structure knowledge and facts stored in so-called triples (\textit{head}, \textit{relation}, \textit{tail}). These triples are used in downstream information extraction tasks such as entity linking, relation extraction, and question answering~\cite{jaradeh2020Question,Cui2017KBQA}. Although large-scale KGs vary in terms of scope and coverage, they are often suffering from incompleteness and sparseness~\cite{Ji2016Knowledge}. Therefore, KGs are required to be regularly updated and completed with missing information.
The incompleteness of KGs motivates the knowledge graph completion task comprising several subtasks~\cite{Yao2019kgbert,Cai2018A}: \rNum{1}) \textit{Link Prediction} aims to find missing head/tail entities in triples. \rNum{2}) \textit{Relation Prediction} predicts  missing relations between two entities. 
KG completion methods calculate the plausibility of a triple via a scoring function to determine the validity of a knowledge graph triple~\cite{Lin2015Learning}. These approaches can be broadly categorized into Knowledge Graph Embeddings (KGE) and Language Models (LM)~\cite{Wang2017Knowledge}.
KGE techniques such as TransE~\cite{Bordes2013Translating} and ConvE~\cite{dettmers2018convolutional} learn entity and relation representation in a low dimensional vector space. However, due to their limitation of utilizing semantic information from text these methods produce different representations for the same entities in distinct triples~\cite{an2018accurate}.
Thus, LM techniques for KG completion emerged in an attempt to represent semantic information encoded in texts and produce contextualized embeddings~\cite{devlin2019bert}.

\textit{\textbf{Motivation.}}
While KGs are a known solution for representing and managing for encyclopedic and common sense knowledge, e.g., in DBpedia~\cite{auer2007dbpedia} and
Wikidata~\cite{Vrande2014Wikidata}, the usage of KGs for scholarly knowledge is a rather new approach. Recently, researchers have focused on building scholarly KGs such as the Open Research Knowledge Graph~\cite{Jaradeh2019Open} and MAG~\cite{farber2019microsoft} to mitigate these previously mentioned challenges and to adhere to the FAIR data principles~\cite{wilkinson2016fair}. In contrast to the typical usage of KGs for encyclopedic and common sense knowledge, managing scholarly knowledge is significantly more challenging due to the heterogeneity, in-homogeneity and evolution of scholarly communication~\cite{hey2009fourth}.
Scholarly KGs differ from encyclopedic purpose KGs because the encoded information is derived from the scholarly literature~\cite{Dessi2020AIKG}.
As a result, the scholarly KGs are ordinarily sparser than generic KGs because structured scholarly knowledge is more laborious to extract~\cite{Jaradeh2019Open}.
Furthermore, these KGs are ambiguous due to a lack of standard terminology used across the literature and poses domain-specific challenges for KG completion task.
Due to domain-specific challenges, the performance of existing KG completion methods such as KG-BERT and TransE are limited when applied to scholarly KG completion tasks (
Section~\ref{sec:eval}).
The observed behavior is not surprising and is due to the peculiar entity and relation types of scholarly KGs.
Hence, we suggest that existing KG completion approaches require additional task-specific context for the scholarly domain.
We focus on the task of KG completion for the scholarly domain.
Inspired by recent advancements in contextualizing language models~\cite{mulang2020evaluating}, we argue that LMs can be utilized for scholarly KG completion tasks if fed with context derived from the scholarly literature.
Our rationale for the choice is as follows: Language models such as SciBERT~\cite{Beltagy2019Scibert} are already trained in an unsupervised manner on large corpora of scholarly literature.
Hence, adapting SciBERT for KG completion will allow us to inherit task-specific context. 
As such, we investigate the following research question:
\vspace{-9pt}
\begin{mdframed}
	\RQN{1}{What is the impact of task-specific context on scholarly KG completion?}
\end{mdframed}

\textit{\textbf{Approach and Contribution.}}
We model KG completion as a sequence classification problem by treating triples of the scholarly KG as sequences of natural language text.
We propose exBERT, a contextualized approach for scholarly KG completion.
In exBERT, we build on SciBERT~\cite{Beltagy2019Scibert} and adapt the underlying pre-trained language model on augmented sequences - augmented by appending types and synonyms of labels for triple elements - for predicting the plausibility of a triple, relation, or an entity.
Our empirical results on the scholarly KG completion datasets provide superior performance on all three KG completion sub-tasks.

\begin{mdframed}
We provide the following three key contributions:
	\begin{enumerate}[leftmargin=0.18cm]
	   	\item[\rNum{1})] exBERT a system that performs scholarly knowledge graph completion task, including the subtasks of triple classification, relation prediction, and link prediction.
		
		\item[\rNum{2})] Extensive evaluation on several datasets to show that proposed method achieves better results against various baselines for the respective tasks. Our proposed datasets, and code are publicly available\footnotemark[1].
		
		\item[\rNum{3})] We release two publicly available scholarly datasets\footnotemark[1] curated from scholarly KGs for the KG completion task. 
	\end{enumerate}
\end{mdframed}
\footnotetext[1]{\url{https://github.com/YaserJaradeh/exBERT}}



\begin{figure*}[ht]
    \centering
    \includegraphics[width=.88\textwidth]{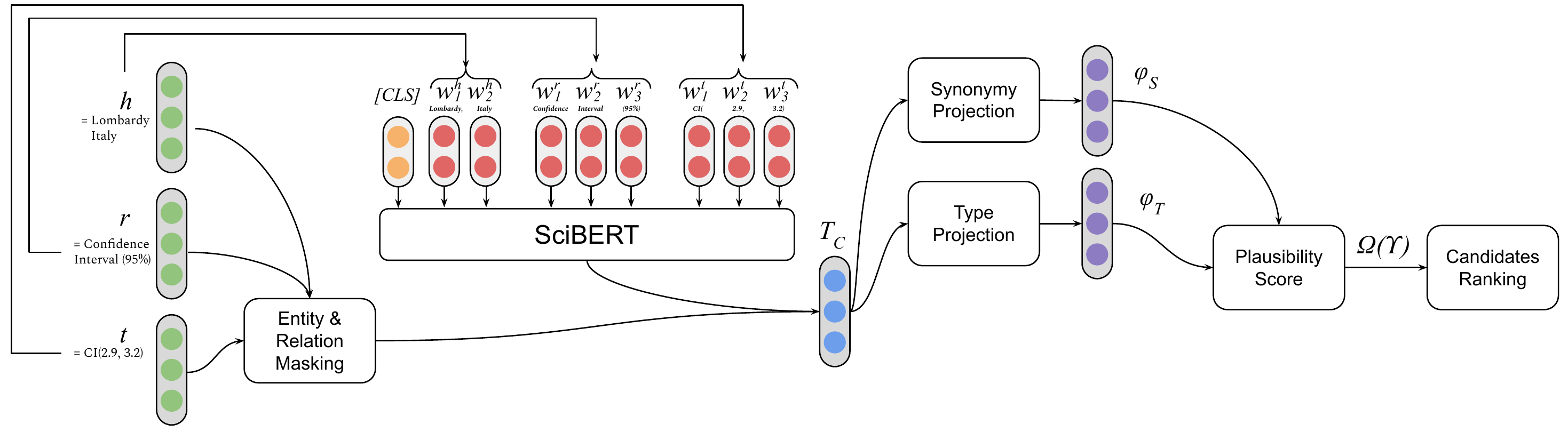}
    \caption{Leveraging triple classification to perform relation and link prediction tasks. $T_{C}$ is the representation of the triple for the classification task. $\varphi_{S}$ is the numerical representation after synonymy projection for relations. $\varphi_{T}$ is the numerical representation after type projection for entities. $\Omega(\Upsilon)$ is the plausibility score of a triple .}
    \label{fig:flow}
\end{figure*}

\section{Related Work}
\label{sec:related}
The KG completion task utilizes different methods and techniques, which can be divided into the two main categories KGE and LM.

\textbf{Knowledge Graph Embeddings (KGE)}:
Knowledge graph embeddings are classified into several categories: translation, semantic matching, and neural network-based models~\cite{Cai2018A}. 
Translational models use distance-based scoring functions to assess the plausibility of a triple $(h,r,t)$. Such models evaluate the plausibility by the distance between the head entity (\textit{h}) and the tail entity (\textit{t}), which is typically done by performing a translation operation by the vector (\textit{r})~\cite{Wang2017Knowledge}. 
\citet{Bordes2013Translating} proposed TransE as a representative model of translational graph embeddings; it uses a negative translational distance function for scoring. Work in \cite{nayyeri2019soft} modified TransE for adapting it in scholarly domain. TransR~\cite{Lin2015Learning} builds entities and relations embeddings by projecting them into different spaces, then building translations. 
Other models rely on semantic matching, which internally uses similarity-based scoring functions. RESCAL~\cite{Nickel2011A}, DistMult~\cite{Yang2014Embedding}, and all their extensions are representatives of this category. Such models employ different scoring functions. For instance, DistMult uses a bilinear function $f(h,r,t)=\left \langle h,r,t \right \rangle$. These methods conduct KG completion using the structural information from the triples disregarding other external information, in particular entity types, logical rules, or textual descriptions. For example, 
\citet{Socher2013Reasoning} represented entities by averaging word embeddings extracted from their labels. 
Compared with these methods, our approach can learn context-aware text embeddings with  pre-trained language models.

\textbf{Language Models (LM):}
Language models that represent nodes and words can be split into two classes: feature-based and fine-tuning approaches. Widely used word embedding techniques such as Word2Vec~\cite{Mikolov2013Distributed} and its extensions aim at adopting features to learn context-independent word embeddings. GloVe~\cite{Pennington2014Glove} is another technique that strives to find a global representation for words based on specific features. FastText~\cite{bojanowski2016enriching} is a technique that uses Skipgram or Bag of Words to compute out-of-vocabulary word vector embeddings. More recently, Flair~\cite{akbik2018coling} embeddings try to generalize embeddings to be more context-aware. 
On the contrary, instead of relying on feature engineering, fine-tuning methods use the model architecture and parameters as a starting point for specific downstream NLP tasks. BERT~\cite{devlin2019bert} and Pegasus~\cite{Zhang2020PEGASUS} are a few examples of pre-trained models that can be fine-tuned on a variety of NLP sub-tasks (\eg Classification, or Summarization). Such models capture the deep semantic patterns in the natural language text and handle text ambiguity and variations. Pre-trained language models have also been used on knowledge graphs, where graph triples are converted into sentences using random walks and then used to train language models~\cite{He2015Learning}. 
Other approaches aim to enhance BERT's representations with entity semantics~\cite{Zhang2019Ernie}. As such, these approaches concentrate on generating new entities and relations. Finally, KG-BERT~\cite{Yao2019kgbert} is a novel approach that utilizes a BERT transformer model to compute plausibility scores of triples based on names or descriptions of entities and relations.

\section{Approach}
\label{sec:approach}
We model the KG completion task as a sequence classification task to harness the richness and power of transformer language models. We rely on a BERT model to perform the classification via transforming the input knowledge graph triples into sequences of text with some extra tokens following the convention of BERT fine-tuning. Furthermore, we leverage a SciBERT model rather than a plain BERT as our core transformer model for the scholarly context.


\subsection{Language Model for Scientific Text}
SciBERT~\cite{Beltagy2019Scibert} is a state-of-the-art pre-trained language model that creates contextual representations using a multi-layer bidirectional transformer encoder architecture described by~\citet{devlin2019bert}. SciBERT builds on the BERT~\cite{devlin2019bert} model and further trains it on scientific literature using 1.14 million scientific papers from Semantic Scholar~\cite{Ammar2018Construction}. The model also consists of 3.17 billion tokens (\ie words). 
SciBERT is pre-trained on two tasks: masked language modeling and next sentence prediction. For next sentence prediction, the language model predicts whether two sequences of input (\ie sentences) are consecutive in the text. For masked language prediction, SciBERT predicts suitably masked input tokens. Furthermore, regarding the fine-tuning phase, SciBERT is initialized with the weights and the parameters from the pre-training phase. Thus, these parameters are fine-tuned using labeled data from downstream natural language processing tasks (\eg question answering, summarization, and token classification)~\cite{Zhang2020Task}.

\subsection{Scholarly Model (exBERT)}

In exBERT, we represent entities and relations of the knowledge graphs using their respective text labels.
Furthermore, we augment these labels with type and synonymy projections. Types of entities are added to their labels and the synonyms of relations are included in the labels as well.
These label sequences are given as an input sentence to our model for further fine-tuning. 
In order to estimate the plausibility score of triples, we arranged the sentences of (\textit{h},\textit{r},\textit{t}) as individual sequences.
In our case, a sequence represents a BERT compatible token sequence combined from two entities (\textit{head} and \textit{tail}) or a complete triple (\textit{head}, \textit{relation}, \textit{tail}).

We extend the simple triples with extra information (\ie types and label synonyms) to further leverage the KGs. We append to the triple the type label of each entity in the head or tail position, and augment the relation representation with label synonyms.

The workflow of exBERT is shown in \autoref{fig:flow}.
exBERT performs a triple classification task to determine if a triple belongs to a KG.
Furthermore, other tasks can be performed by leveraging the triple classification task, \eg head/tail or relation prediction.
The first token of every input sequence for exBERT is always a unique classification token \texttt{[CLS]}.
Each entity in the triple and the relation are represented as a sentence containing a list of tokens $\nu_{1}, ..., \nu_{\alpha}$, $\alpha \ge 1$.
For instance, \emph{Lombardy Italy has a confidence interval (95\%) of CI(2.9, 3.2)} has a head entity \emph{Lombardy, Italy} which comprises two tokens $\nu_{1}^{h}$ = Lombardy, $\nu_{2}^{h}$ = Italy; the relation \emph{Confidence Interval (95\%)} comprises three tokens $\nu_{1}^{r}, \nu_{2}^{r}, \nu_{3}^{r}$; and the tail \emph{CI(2.9, 3.2)} comprises three tokens $\nu_{1}^{t}, \nu_{2}^{t}, \nu_{3}^{t}$.
While constructing the sentences of entities and relations, a special token \texttt{[SEP]} is used to differentiate elements (\ie components of the triple).
The separation token indicates that the various elements of the sentence have different segment embeddings.
However, the tokens of the head and tail entities share identical segment embeddings.
Furthermore, when various tokens occupy the same position -- for the case of SciBERT, a position $1 \leq i \leq 512$ -- they have the same position embeddings. 

Token sequences are used as an input to the SciBERT model architecture, which is a multi-layer bidirectional transformer encoder based on the native BERT architecture~\cite{devlin2019bert}. The final hidden vector of the special \texttt{[CLS]} token is denoted as $C \in \mathbb{R}^{H}$, where $H$ is the hidden state size in the pre-trained SciBERT model. Furthermore, the $i$-th token of the model's input tokens is referred to as $\nu_{i} \in \mathbb{R}^{H}$. When fine-tuning the model to perform the triple classification task, a set of weights is created (\ie classification layer weights) $W \in \mathbb{R}^{2 H}$. A sigmoid function $\Omega$ is used to score a triple $\Upsilon = (h, r, t)$ and produce its class affiliation.
$\Omega$ is defined as $\Omega^{\Upsilon} = sigmoid(CW^{\nu})$.
Where $\Omega^{\Upsilon} \in \mathbb{R}^{2}$ is a two-dimensional real-valued vector with $\Omega^{\Upsilon}_{0}, \Omega^{\Upsilon}_{1} \in \left [ 0, 1 \right ]$ and $\Omega^{\Upsilon}_{0} + \Omega^{\Upsilon}_{1}=1$.

We are then able to compute the cross-entropy loss using the scoring function $\Omega^{\Upsilon}$ and predicted labels $y^{\Upsilon}$ as follows:

\begin{equation}
   \Phi  = - \sum_{\Upsilon \in \mathbb{K^{+}} \cup \mathbb{K^{-}}} \left ( y^{\Upsilon}log(\Omega^{\Upsilon}_{0}) + (1-y^{\Upsilon})log(\Omega^{\Upsilon}_{1}) \right )
    \label{eq:cross-entropy}
\end{equation}

\noindent
where $\mathbb{K^{+}}$ is the positive triple set and $\mathbb{K^{-}}$ is the negative triple set. $y^{\Upsilon}$ is the label of the triple (\ie positive or negative label) $y^{\Upsilon} \in \left \{ 0,1 \right \}$. While the positive triple set contains the correct triples within the KG, the negative set $\mathbb{K^{-}}$ is constructed by replacing the head entity $h$ or tail entity $t$ in a positive triple with a random entity $\bar{h}$ or $\bar{t}$.




When corrupting the triple (\ie creating the negative set), we make sure that the head or tail being replaced is not the correct one and ensure that the resulting corrupt triple does not belong to the positive set. Both relation and link prediction tasks use triple classification as an underlying task. The difference is in the way these tasks compose the input sentences (using only entities or the complete triple). Based on the findings of~\citet{Yao2019kgbert}, we compose the input sequence with the two entities $h$ and $t$ to predict a relation~$r$. KG-BERT empirical results suggest that predicting relations from the two entities using triple classification has higher performance. The other setting involving complete triples by curating negative samples with random relations $\bar{r}$ does not yield a performance gain. Similarly to link prediction, the final hidden state vector $C$ corresponds to the special classification token \texttt{[CLS]}.
We highlight that relation prediction exBERT differs from link prediction exBERT (\ie head and tail prediction) with the classification layer weights.
For relation prediction, the tasks fine-tune the weights $\bar{W} \in \mathbb{R}^{R \times H}$, whereby $R$ is the number of all KG relations. The scoring function is $\bar{\Omega}^{\Upsilon} = softmax(C\bar{W}^{\nu})$, whereby $\bar{\Omega}^{\Upsilon} \in \mathbb{R}^{R}$ is a \textbf{R}-dimensional real vector with $\bar{\Omega}^{\Upsilon}_{i} \in \left [ 0, 1 \right ]$ and $\sum_{i}^{R} \bar{\Omega}^{\Upsilon}_{i} = 1$. Similarly, we compute the cross-entropy loss with the help of the scoring function $\bar{\Omega}^{\Upsilon}$ and the relation labels $\bar{y}^{\Upsilon}$ as shown in \autoref{eq:cross-entropy-relation}.

\begin{equation}
\small
    \Phi_{r}  = - \sum_{\Upsilon  \in \mathbb{K^{+}}} \sum_{i=1}^{R} \bar{y}^{\Upsilon}_{i} log(\bar{\Omega}^{\Upsilon}_{i}),\textup{ } \bar{y}^{\Upsilon}_{i} = \begin{cases}
1 & \text{ if }  r=i \\
0 & \text{ o/w }  
\end{cases}
    \label{eq:cross-entropy-relation}
\end{equation}

\noindent
where $\bar{y}^{\Upsilon}_{i}$ is the relation class (\ie indicator) for a positive triple $\Upsilon$ and its value is conditional on the relation.

\begin{table}[!b]
\centering
\caption{Statistics of datasets. We propose two new datasets (labeled with [$^{\star}$]) for the scholarly KG completion task.}
\resizebox{.9\columnwidth}{!}{
\renewcommand{\arraystretch}{1.2}
\begin{tabular}{c|cccccc}
\hline
\bf{Dataset}& \bf{\# Ent} & \bf{\# Rel}	& \bf{\# Triples}& \bf{\# Train}& \bf{\# Dev} & \bf{\# Test} \\
\hline
\textit{ORKG21*} & 226,210 & 2634 & 249,682 & 149,808 & 49,937  & 49,937  \\ 
\textit{PWC21*} & 192,115  & 26 & 284,875 & 170,925 & 56,975 & 56,975  \\
\textit{UMLS~\cite{Bodenreider2004UMLS}} & 135 & 46 & 6,529 & 5,216 & 652 & 661  \\
\hline
\end{tabular}
}
\label{tab:datasets-stat}
\end{table}

\section{Evaluation}
\label{sec:eval}
We conduct our experiments and analysis in response to the overall research question \RQN{1} (see \autoref{sec:intro}). As such, we also compare exBERT against approaches that do not come from a scholarly context. To understand the efficacy of exBERT for scholarly KG completion, we further divide our overall research question into three sub-questions: \RQN{1.1}{What is the performance of exBERT for scholarly relation prediction?}, \RQN{1.2}{What is the performance of exBERT for link prediction in scholarly KGs?}, and \RQN{1.3}{What is the efficacy of exBERT in scholarly triple classification?}

\paragraph{\textbf{Datasets}} 
The scholarly KG domain is relatively new~\cite{Jaradeh2019Open}. There is a scarcity of standard datasets to benchmark the performance of KG completion methods. 
Therefore, we created two datasets collected from available knowledge graphs and online resources.

\rNum{1}) \textbf{ORKG21}: A dataset of scholarly contributions extracted from the Open Research Knowledge Graph (ORKG) infrastructure~\cite{Jaradeh2019Open}. The ORKG contains data on scientific contributions and publications curated by crowdsourcing that is complemented by automatic processes. We created the dataset based on a data dump provided by the ORKG in RDF~\cite{manola2004rdf} format.
The resulting dataset contains many relationships because ORKG relies primarily on crowdsourcing for data entry and does not automate end-to-end extraction. As a consequence, multiple relationships are created by the different crowd members, but represent similar intentions, resulting in a significant number of relationships with slightly different representations.

\rNum{2}) \textbf{PWC21}: A dataset from the online resource Papers-with-Code that describes papers in the field of machine learning, information extraction, and NLP along with their evaluation results. The PWC data is represented in tabular rather than graph structure. We transformed the raw data into RDF for broader use. The resulting dataset contains only a small set of relations because it focuses on certain aspects of research papers (\ie evaluation results and metrics).

Since both datasets contain literals, we transformed them into entities by creating sequential IDs in the form of "/literal\_\textbf{num}", similar to the Yago3-10 dataset~\cite{sun2018rotate}. In addition to the two datasets we created, we also used the public scientific dataset \textbf{UMLS}~\cite{Bodenreider2004UMLS}. The  \qq{Unified Medical Language System} is an ontology for the medical field that describes technical medical concepts and their interlinked relations.
\autoref{tab:datasets-stat} provides a summary of the employed datasets. If a dataset does not provide a class label (\ie binary class) for training/testing triples, throughout our evaluation we considered every triple in the dataset as ground truth, and we corrupted triples to generate negative samples as explained in \autoref{sec:approach}.

\paragraph{\textbf{Baselines}} 
We compare exBERT with several state-of-the-art KGE methods to evaluate its efficacy with respect to the tasks:

\rNum{1}) TransE~\cite{Bordes2013Translating} and its extensions TransH~\cite{Wang2014TranH}, TransD~\cite{ji2015Knowledge}, TransR \cite{Lin2015Learning}, TransG~\cite{Xiao2016TransG}, and DistMult~\cite{Yang2014Embedding}, which only rely on structural information of the knowledge graph to compute the embeddings.
\rNum{2}) NTN~\cite{Socher2013Reasoning} and its simplified version ProjE~\cite{Shi2017ProjE}.
\rNum{3}) ConvKB~\cite{Nguyen2018Novel} which is a CNN model.
\rNum{4}) Other KGE models, specifically: AATE~\cite{an2018accurate} and TEKE~\cite{wang2016text} that leverage textual information; Contextualized graph embeddings with KG2E~\cite{He2015Learning}; Complex-values KG embeddings techniques RotatE~\cite{sun2018rotate} and ComplEX~\cite{Trouillon2016Complex}; compositional vector-space embeddings HolE~\cite{Nickel2016Holographic}; Learning embeddings dependently with SimplE~\cite{Kazemi2018SimplE};
\rNum{5}) BERT based approach KG-BERT by~\citet{Yao2019kgbert} which utilizes a BERT-Base model to perform KG completion.

It is noteworthy that the baselines vary by task because some KGE models do not perform tasks such as relation prediction. 
\vspace{-2pt}
\paragraph{\textbf{Experimental Settings}}
SciBERT is the base transformer model used in exBERT.
It has $12$ layers, $12$ self-attention heads, and $H = 768$ hidden layers.
We used the Adam implementation for the optimizer.
For fine-tuning the triple classification task, the batch size is $32$, with a learning rate $5e-5$ and a dropout rate of $0.1$.
For triple classification, we sample one negative triple for a positive triple to assure class balance for binary classification tasks.
Furthermore, the number of epochs for triple classification is $3$.
We found no discernible improvement by increasing this number. For the link prediction task, we used $5$ epochs, and we also sampled $5$ negative samples for each positive triple (following the findings of~\citet{Yao2019kgbert}). Finally, for relation prediction, the number of epochs chosen was $20$.
For benchmarking exBERT and KG-BERT, we used a system running Ubuntu 20.04 with 128GB of RAM and 4x Nvidia A100 GPUs, each with 40GB vRAM. Other KGE baselines are trained on 8x Nvidia RTX 3090 GPUs, each with 24GB vRAM. KG embeddings are benchmarked using PyKeen~\cite{Ali2021PyKeen}. 

\vspace{-2pt}
\paragraph{\textbf{Evaluation Metrics}} 
Following the widely adapted metrics and inheriting evaluation settings from KG-BERT~\cite{Yao2019kgbert}, we report the Mean Rank (MR) and the cut-off hit ratio (Hits@N) metrics on all the datasets for the link prediction ($N=10$) and the relation prediction ($N=1$) tasks. 
MR reports the average rank of all correct entities. Hits@N evaluates the ratio of correct entity predictions at a top N predicted results. Similar to KG-BERT, for the triple classification task we report accuracy. Classification accuracy is the number of correctly classified triples of the testing set divided by the total number of test triples.
For the relation prediction task, we rank candidate relations by the scoring function $f(h,r,t)$ (\cf $\bar{\Omega}^{\Upsilon}$ in our approach formalization). Each correct test triple $(h,r,t)$ is corrupted by replacing the relation with every other relation in the KG, with the exception of the relation itself, \ie $r' \in R | r' \neq r$. Then these candidates are ranked in descending order by their plausibility score we obtain from the triple classification task. Following \citet{Yao2019kgbert} and \citet{Bordes2013Translating} we report the results under the filtered settings only, which means that all corrupted triples are removed from training, development, and testing sets before getting the ranking lists.
In the link prediction task, we aim to predict the missing entity of the triple. Each correct test triple $T=(h,r,t)$ is corrupted by replacing either the head entity or the tail entity with every other entity in the KG $e \in E | e \neq h \wedge e \neq t$.

\subsection{Experiment 1: Relation Prediction}
Table~\ref{tab:relation-prediction} presents relation prediction results on all three datasets. 
Structural embeddings such as TransE, TransH, and TransR report limited performance due to the lack of contextualized information. KG2E and NTN perform slightly better than TransE and extensions with their density-based embeddings and neural model, respectively. PairRE outperforms the other embedding-based models with its paired encoding of relations and its capability of encoding various types of relations (\eg symmetric, inverse). A common downfall of embedding-based techniques is that they learn identical embedding representations of entities and relations and do not account for the different meanings that words might have in various contexts. The limited performance in the scholarly domain clearly validates the observation. In contrary,
KG-BERT and exBERT significantly outperform embedding-based models due to their own contextualized embeddings learned using a large corpus of unstructured text. However, task-specific context enhanced the ability of exBERT in predicting scholarly relations compared to KG-BERT, which successfully answers our first sub-question \textbf{RQ1.1}.

\begin{table}[t]
\centering
    \caption{Relation prediction results on all three datasets. Best results are indicated in bold font. The results listed in the table were all obtained by us. Techniques marked by [$^{\star}$] do not perform relation prediction by default. Hence, we employ triple classification to measure performance for relation prediction.}
\resizebox{.9\columnwidth}{!}{
\begin{tabular}{l|cc|cc|cc}
    \hline
    \multirow{2}*{\textbf{Method}} &\multicolumn{2}{c|}{\textbf{ORKG21}} & \multicolumn{2}{c|}{\textbf{PWC21}} & \multicolumn{2}{c}{\textbf{UMLS}}\\
    \cline{2-7}
    & MR & Hits@1  & MR & Hits@1 & MR & Hits@1\\
    \hline
    TransE~\cite{Bordes2013Translating}& 1683 & 50.1 & 1232 & 49.5 & 166 & 23.3 \\
    TransH~\cite{Wang2014TranH}& 1544 & 51.8 & 1169 & 51.1 & 141 & 25.7 \\
    TransR~\cite{Lin2015Learning}& 1378 & 53.3 & 897 & 54.4 & 78 & 27.3 \\
    NTN$^{\star}$~\cite{Socher2013Reasoning}& 1189 & 58.4 & 754 & 59.2 & 22 & 29.9 \\
    KG2E$^{\star}$~\cite{He2015Learning}& 1301 & 57.9 & 773 & 58.3 & 21 & 30.0 \\
    ProjE~\cite{Shi2017ProjE}& 522 & 74.6 & 212 & 78.5 & 6.42 & 41.8 \\
    PairRE~\cite{Chao2021PairRE}& 206 & 82.1 & 59 & 88.6 & 4.25 & 63.5 \\
    KG-BERT~\cite{Yao2019kgbert}& 15.37 & 92.8 & 1.51 & 96.7 & 1.21 & 87.2 \\
    exBERT& \textbf{12.98} & \textbf{95.5} & \textbf{1.02} & \textbf{98.3} & \textbf{1.11} & \textbf{88.8} \\
    \hline
    \end{tabular}
}
\label{tab:relation-prediction}
\end{table}

\subsection{Experiment 2: Link Prediction}
We report link prediction results in \autoref{tab:link-prediction}. Across all datasets, exBERT achieves significantly higher performance compared to all baselines. On the ORKG21 dataset, the majority of relations are symmetric relations. Translation-based methods report limited performance on ORKG21 due to their inability to infer symmetric connectivity patterns of a KG. Furthermore, the TransX family uses only the structure of the KG and does not induce context information or labels of entities, which results in limited performance.
On PWC21, connectivity patterns are evenly distributed across all relation types (one-to-one, many-to-many, etc.), and the performance of exBERT does not drop. The majority of entities contain multiple relations on the UMLS dataset, and exBERT can successfully predict the missing links while reporting a slightly lower performance compared to best results. Embedding-based models continued to show limited performance for link prediction tasks due to non-standard characteristics of scholarly entities. 
From our empirical results, we conclude that the task-specific context fed into exBERT for the scholarly domain has positively impacted the performance across all KG completion tasks (successfully answering \textbf{RQ1.2}).

\begin{table}[t]
\centering
\caption{Link prediction results on all three datasets. Best results are indicated in bold font. Results accompanied by asterisk$^{*}$ are reported by ~\citet{Yao2019kgbert}, other results were obtained by us and not from respective publications.}
\resizebox{.9\columnwidth}{!}{
\begin{tabular}{l|cc|cc|cc}
    \hline
    \multirow{2}*{\textbf{Method}} &\multicolumn{2}{c|}{\textbf{ORKG21}} & \multicolumn{2}{c|}{\textbf{PWC21}} & \multicolumn{2}{c}{\textbf{UMLS}}\\
    \cline{2-7}
    & MR & Hits@10  & MR & Hits@10 & MR & Hits@10\\
    \hline
    TransE~\cite{Bordes2013Translating} & 2879 & 51.2 & 3176 & 60.8 & 1.84$^{*}$ & 89.9$^{*}$ \\   
    TransH~\cite{Wang2014TranH} & 2811 & 52.5 & 2994 & 61.3 & 1.80$^{*}$ & \textbf{99.5}$^{*}$ \\
    TransD~\cite{ji2015Knowledge} & 2791 & 53.2 & 2135 & 68.8 & 1.71$^{*}$ & 99.3$^{*}$ \\ 
    ConvKB~\cite{Nguyen2018Novel} & 216 & 70.1 & 388 & 72.9 & - & -   \\
    ComplEX~\cite{Trouillon2016Complex} & 713 & 65.3 & 456 & 72.7 & 2.59$^{*}$ & 96.7$^{*}$   \\
    HolE~\cite{Nickel2016Holographic} & 98 & 73.7 & 97 & 76.4 & -  & -   \\
    CompGCN~\cite{Vashishth2020CompositionBased} & 2.84 & 84.6 & 4.02 & 82.7 & - & -   \\
    SimplE~\cite{Kazemi2018SimplE} & 3.40 & 82.4 & 3.91 & 82.9 & - & -   \\
    KG-BERT~\cite{Yao2019kgbert}& 2.03 & 86.1 & 4.03 & 82.7 & \textbf{1.47}$^{*}$ & 99.0$^{*}$ \\
    exBERT& \textbf{1.80} & \textbf{87.4} & \textbf{2.11} & \textbf{84.2} & 1.97 & 98.9 \\
    \hline
\end{tabular}
}
\label{tab:link-prediction}
\end{table}

\begin{table}[bp]
    \centering
    \caption{Triple classification accuracy (in percent) for different embedding methods. The listed results were obtained by us and are not from the corresponding publications. Best results are indicated with bold font.}
    \resizebox{.75\columnwidth}{!}{
    \renewcommand{\arraystretch}{1.1}
    \begin{tabular}{l|ccc|c}
    \hline
    \bf{Method}& \textbf{ORKG21} & \textbf{PWC21} & \textbf{UMLS} & \textbf{Avg.} \\
    \hline
    TransE~\cite{Bordes2013Translating} & 77.6 & 77.3 & 78.1 & 77.7 \\   
    TransH~\cite{Wang2014TranH} & 78.8 & 77.9 & 79.2 & 78.7 \\
    TransR~\cite{Lin2015Learning} & 81.4 & 81.3 & 81.9 & 81.5 \\
    TransD~\cite{ji2015Knowledge} & 84.3 & 83.6 & 84.9 & 84.2 \\ 
    TransG~\cite{Xiao2016TransG} & 85.1 & 84.7 & 85.2 & 85.0  \\
    TEKE~\cite{wang2016text} & 84.8 & 84.2 & 84.9 & 84.6 \\ 
    KG2E~\cite{He2015Learning} & 79.6 & 78.8 & 79.7 & 79.4 \\
    DistMult~\cite{Yang2014Embedding} & 86.2 & 86.2 & 86.8 & 86.4 \\
    ConvKB~\cite{Nguyen2018Novel} & 87.3 & 86.8 & 83.1 & 85.7   \\
    HolE~\cite{Nickel2016Holographic} & 87.3 & 87.3 & 88.2 & 87.6   \\
    NTN~\cite{Socher2013Reasoning} & 85.0 & 84.9 & 85.1 & 85.0   \\
    SimplE~\cite{Kazemi2018SimplE} & 89.7 & 89.2 & 89.1 & 89.3   \\
    KG-BERT~\cite{Yao2019kgbert} & 95.1 & 93.3 & 89.7 & 92.7 \\
    exBERT & \textbf{97.1} & \textbf{96.0} & \textbf{90.3} & \textbf{94.5} \\
    \hline
    \end{tabular}
    }
    \label{tab:triple-classification}
\end{table}

\begin{figure*}
    \begin{subfigure}{\textwidth}
        \centering
        \includegraphics[width=.9\textwidth]{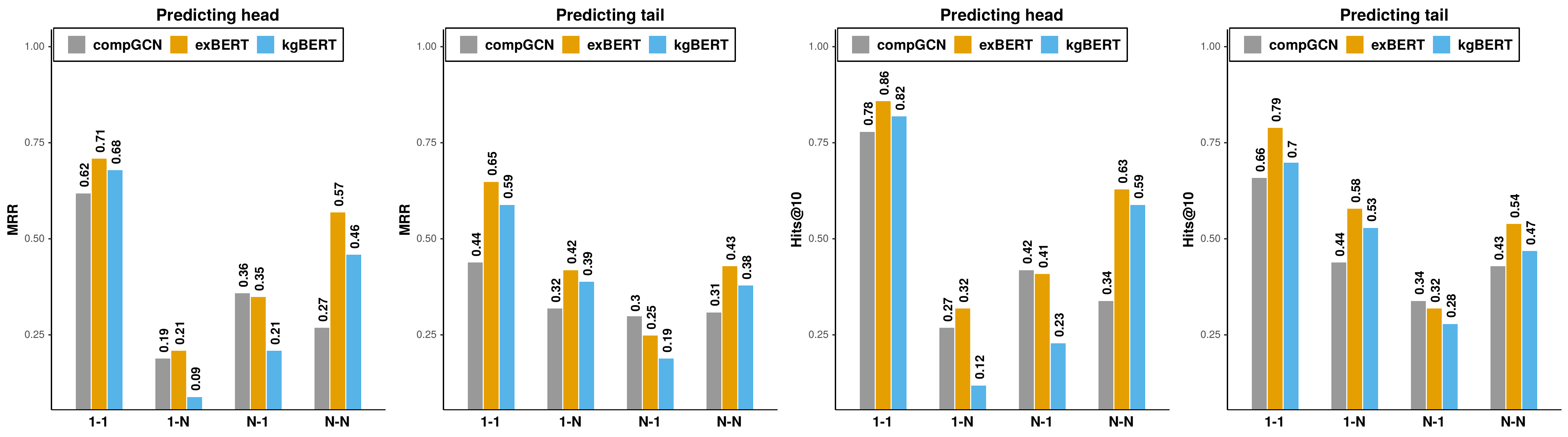}  
        \caption{Link prediction performance on ORKG21.}
        \label{fig:link-orkg}
    \end{subfigure}
    \begin{subfigure}{\textwidth}
        \centering
        \includegraphics[width=.9\textwidth]{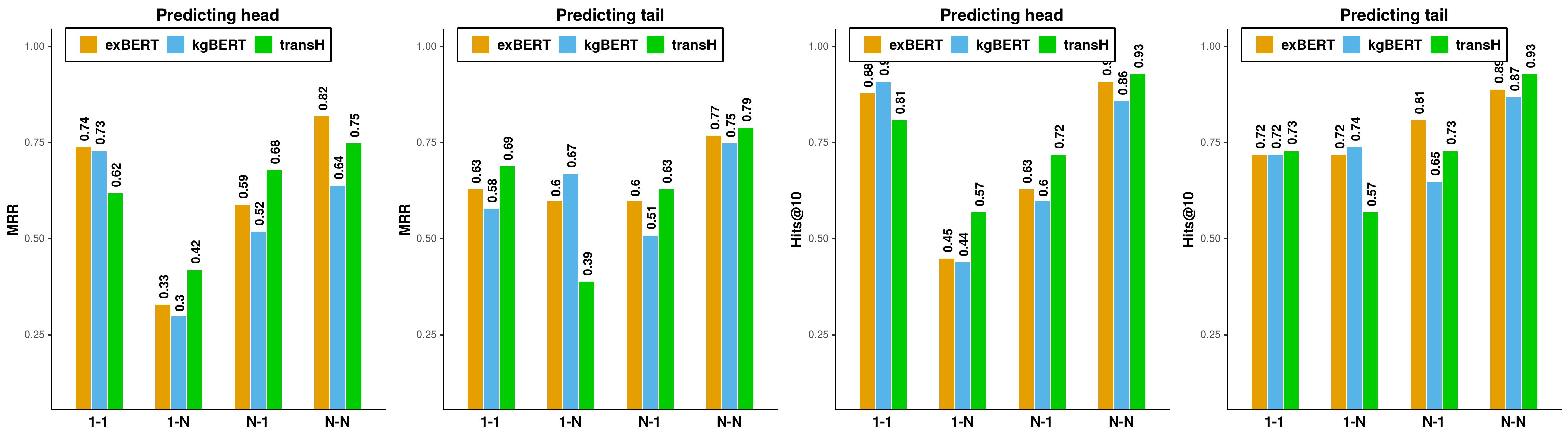}  
        \caption{Link prediction performance on UMLS.}
        \label{fig:link-umls}
    \end{subfigure}
    \begin{subfigure}{\textwidth}
        \centering
        \includegraphics[width=.9\textwidth]{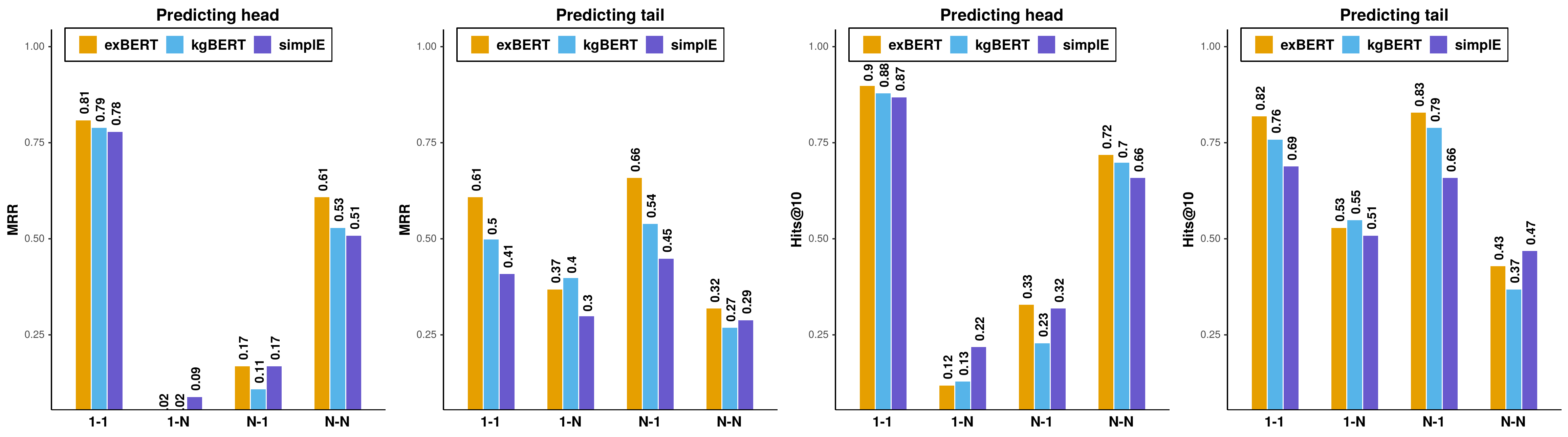}  
        \caption{Link prediction performance on PWC21.}
        \label{fig:link-pwc}
    \end{subfigure}
    \caption{Link prediction performance for the top three performing baselines per dataset against exBERT for each relation type. We show two metrics MRR and Hits@10 for each relation type (1-1, 1-N, N-1, N-N). We observe that depending on the dataset and sub-task (head/tail prediction), performance varies per relation type. The fine-grained analysis provides a detailed overview of the strength and weakness of exBERT against best performing models.}
\label{fig:ablation-link-prediction}
\end{figure*}

\begin{figure*}
    \centering
    \includegraphics[width=.85\textwidth]{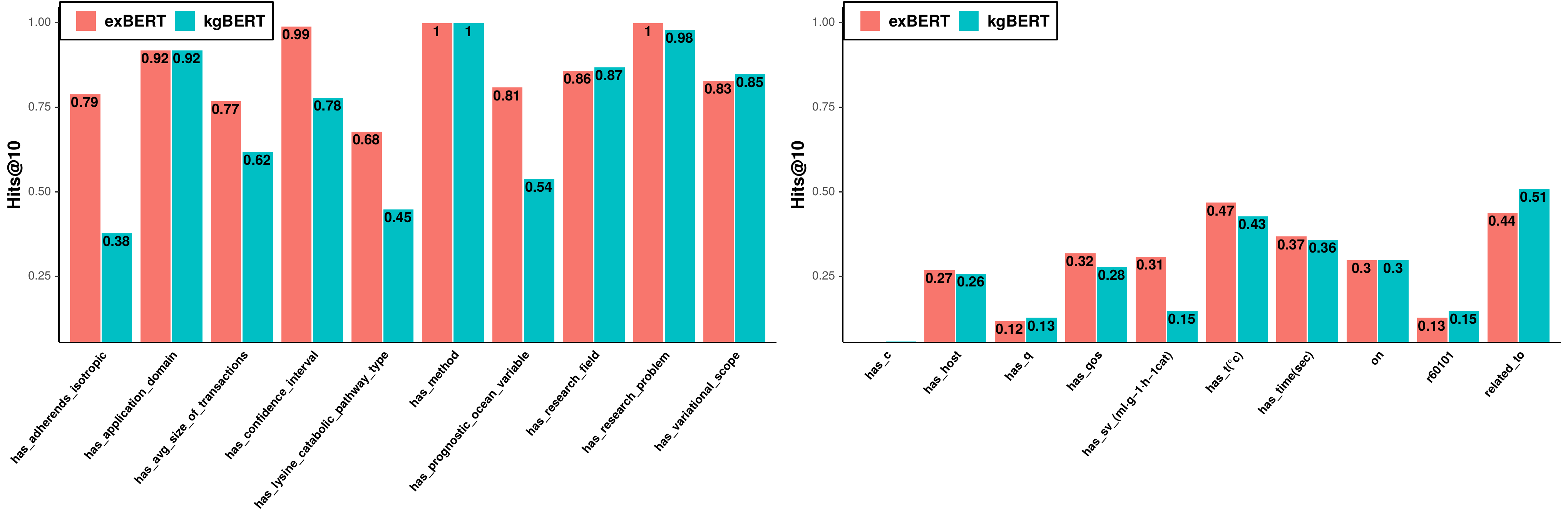} 
    \caption{Illustration of the Hits@10 metric for the ten best (left plot) and worst (right plot) predicted relations in the ORKG21 dataset by exBERT and the second best baseline KG-BERT.}
    \label{fig:ablation-relation-prediction}
\end{figure*}

\subsection{Experiment 3: Triple Classification}
The objective of triple classification is to assign a score to each triple $T=(h,r,t)$ depending on whether or not it belongs to the underlying KG.
\autoref{tab:triple-classification} shows the classification results of exBERT against all other baselines on the benchmark datasets. exBERT outperforms all baseline KGE approaches confirming the effectiveness of our approach for the scholarly domain. We note that translation-based KGE such as TranE could not achieve high scores because of the cardinality of the knowledge graph relations (\ie the existence of one-to-many, many-to-one, and many-to-many relations). However, its extensions, TransH, TransR, TransD, and TransG outperform TransE by introducing relation-specific parameters. The DistMult model performs relatively better than the Translation family. The CNN models, \eg ConvKB, perform well, suggesting that a convolution model can capture global correlations among entities and relations in the scholarly domain. HolE performs proportionately well and comparably to ConvKB. SimplE achieves the best results compared to the other embedding-based approaches. Finally, exBERT outperforms KG-BERT leveraging the underlying semantic and contextualized knowledge of a transformer language model trained on scientific data. 

Based on our observations, we identify the following reasons for the superior performance of exBERT for the triple classification task:
Firstly, the triple classification task is akin to the next sentence prediction, for which SciBERT is pre-trained with large text corpora. The fine-tuning weights are already positioned for the inference of correlation among triple components. And mostly, contextualized embeddings for the scholarly domain are explicit with our approach via the hidden token vectors. This has positively impacted the overall performance (successfully answering \textbf{RQ1.3}, and collectively answering \textbf{RQ1}).


\subsection{Ablation Studies}


\paragraph{\textbf{Impact of relation type on the link prediction task}}
For each dataset, we show the metrics (Mean reciprocal rank (MRR) because it is in range [0,1] and Hits@10) for exBERT and the top-performing baselines for various relation types (\ie one-to-one, one-to-many, many-to-one, many-to-many). \autoref{fig:ablation-link-prediction} presents the performance of this task for every relation type and all datasets. Based on the fine-grained analysis, we suggest that: 
\rNum{1}) On the ORKG21 dataset, exBERT maintains superior performance across all relation types for both metrics. CompGCN performs slightly better than KG-BERT and exBERT on many-to-one relation types. One possible reason is that CompGCN jointly embeds nodes and relations in a graph that permits the model to handle dense relations~\cite{Vashishth2020CompositionBased}.
\rNum{2}) On the UMLS dataset, exBERT suffers a performance drop in the head prediction that results in a lower performance observed in \autoref{tab:link-prediction}.
\rNum{3}) On the PWC21 dataset, it is interesting to observe that all three models suffer significant performance drops for predicting head entities in the one-to-many relation category. However, models maintain steady performance across all other relation categories. 

\paragraph{\textbf{Relation Prediction results for best/worst relations}}
To further comprehend how relations affect the overall performance of the relation prediction task, we select the best/worst ten performing relations in the ORKG21 dataset. We plot the evaluation metrics graphs for exBERT compared to the second best baseline KG-BERT. \autoref{fig:ablation-relation-prediction} illustrates the individual performance. KG-BERT performs comparably to exBERT for the generic relation types such as \textit{has\_method}. However, for peculiar scholarly relations, such as \textit{has\_adrehends\_isotrophic} and \textit{has\_prognostic\_ocean\_variable}, the performance of KG-BERT is limited due to missing task-specific context. Observed results validate our hypothesis to supplement exBERT with scholarly context for the KG completion task. 

In \autoref{fig:ablation-relation-prediction}, most of the relations are quite ambiguous. Furthermore, in some cases, the scholarly context does not positively impact the performance of exBERT compared to KG-BERT. For instance, KG-BERT performs better than exBERT for the relation type \textit{related\_to}. KG-BERT is trained on a large unstructured corpus from generic domains, and \textit{related\_to} is a commonly occurring relation between two real-world entities.

\section{Conclusion and Future Work}
\label{sec:conclusion}
The hypothesis investigated in this paper was to study if task-specific context has an impact on scholarly KG completion task. For the same, we proposed exBERT and provided a set of experiments illustrating the positive impact of scholarly context encoded in exBERT for the KG completion task. We model the KG completion task as a sequence classification task, where we considered each KG triple as a set of sequences in a natural language. This allowed us to utilize SciBERT as an underlying model and adapt it for KG completion in the scholarly domain. We systematically
studied the impact of our choices in the proposed approach. For instance, the
ablation study demonstrates the effectiveness of scholarly context and provides insights on the strengths and weaknesses of exBERT. Albeit effective, exBERT is the first step of a larger research agenda. Based on our observations, we see the following open research questions in this domain:
\rNum{1}) While scholarly context has positively impacted the performance, there are several relation types for which exBERT showed limited performance. One potential reason is the scarcity of training occurrences for certain relations such as \textit{has\_qos}. We believe that for unseen entities and relations, a zero-shot setting would be more suitable. Additional experiments are needed to verify the observation, and we point readers to this open research direction as the next step. 
\rNum{2}) Considering that exBERT relies on labels and textual descriptions of an entity, data quality is a crucial aspect---as is typical for knowledge-intensive tasks. 
How the quality of contextual data impacts the performance of scholarly KG completion approaches is an open research direction.
\rNum{3}) Several recent approaches utilize KGs as an additional source of background knowledge to provide context in downstream tasks such as entity linking and relation extraction. How context derived from the scholarly KGs supplementing the textual context can help in effective KG completion is a further open question.
\rNum{4}) Our work focuses on English and developing multi-lingual KG completion approaches is a viable next step. 

\begin{acks}
This work was co-funded by the European Research Council for the project ScienceGRAPH (Grant agreement ID: 819536) and the TIB Leibniz Information Centre for Science and Technology. We thank Oliver Karras and Allard Oelen for their valuable feedback.
\end{acks}

\bibliographystyle{ACM-Reference-Format}
\bibliography{references}

\end{document}